\documentclass{article}

\usepackage[final]{neurips_2024}

\usepackage[utf8]{inputenc} 
\usepackage[T1]{fontenc}    
\usepackage{hyperref}       
\hypersetup{colorlinks = true, urlcolor = blue, linkcolor = blue, citecolor= blue}
\usepackage{url}            
\usepackage{booktabs}       
\usepackage{amsfonts}       
\usepackage{nicefrac}       
\usepackage{microtype}      
\usepackage{xcolor}         

\usepackage{multirow}
\usepackage{pdflscape}
\usepackage{changepage}
\usepackage{float}
\usepackage{makecell}
\usepackage{amsmath}
\usepackage{graphicx}
\usepackage{tabularx}
\usepackage{titlesec}
\titlespacing*{\subsection}{0pt}{0.5\baselineskip}{0.2\baselineskip} 
\titlespacing*{\section}{0pt}{0.5\baselineskip}{0.2\baselineskip} 
\titlespacing*{\paragraph}{0pt}{0.2\baselineskip}{0.2\baselineskip} 

\title{Improving In-Context Learning with \\ Small Language Model Ensembles}

\author{%
  M. Mehdi Mojarradi\thanks{Correspondence to: Mehdi Mojarradi <m.mehdi.mojarradi@gmail.com>.} \\
  Oxford Internet Institute\\
  University of Oxford\\
  \And
  Lingyi Yang \\
  Mathematical Institute \\
  University of Oxford \\
  \AND
  Robert McCraith \\
  Oxford Internet Institute \\
  University of Oxford \\
  \And
  Adam Mahdi \\
  Oxford Internet Institute \\
  University of Oxford \\
}

\begin{document}

\maketitle

\begin{abstract}
    Large language models (LLMs) have shown impressive capabilities across various tasks, but their performance on domain-specific tasks remains limited. While methods like retrieval augmented generation and fine-tuning can help to address this, they require significant resources. In-context learning (ICL) is a cheap and efficient alternative but cannot match the accuracies of advanced methods. We present Ensemble SuperICL, a novel approach that enhances ICL by leveraging the expertise of multiple fine-tuned small language models (SLMs). Ensemble SuperICL achieves state of the art (SoTA) results on several natural language understanding benchmarks. Additionally, we test it on a medical-domain labelling task and showcase its practicality by using off-the-shelf SLMs fine-tuned on a general language task, achieving superior accuracy in large-scale data labelling compared to all baselines. Finally, we conduct an ablation study and sensitivity analyses to elucidate the underlying mechanism of Ensemble SuperICL. Our research contributes to the growing demand for efficient domain specialisation methods in LLMs, offering a cheap and effective method for practitioners.\footnote[1]{Code: \url{https://github.com/mehdimojarradi/Ensemble-SuperICL}.}

\end{abstract}

\section{Introduction}




In-context learning (ICL) is an effective method for adapting large language models (LLMs) to perform specific tasks without the need of updating model parameters through fine-tuning \citep{brown2020language}. It involves prompting an LLM with few-shot training examples and a test input, allowing the LLM to infer the correct output from the provided context.



While ICL is a time and cost-efficient method, it lacks the accuracy of more expensive methods such as fine-tuning and retrieval-augmented generation (RAG). Addressing this limitation is significant in light of two trends. First, there is a growing need for adapting LLMs to specific tasks or specialised domains, such as question answering on specific corpora (e.g. medical or legal documents) \citep{ling2023domain}. Second, increasing computational and data bottlenecks are creating a growing demand for methods that do not require large compute and data resources, unlike RAG and fine-tuning  \citep{thompson2020computational}. Improving the performance of ICL can provide practitioners with cheap but highly accurate domain specialisation in LLMs.


The need for improved ICL performance has led to a line of studies dedicated to constructing prompts that achieve high and stable ICL performance \citep{gao2020making, liu2021makes, su2022selective, levy2022diverse, ye2023compositional}. In \citet{supericl} and \citet{yang2023supervised}, the authors proposed SuperICL, which appends the predictions of a fine-tuned small language model (SLM)\footnote{Henceforth in this paper, small language model (SLM) refers to a language model with less than 1B parameters that is fine-tuned on a task-specific dataset.} to ICL prompts, combining the specialised knowledge of an expert model with the general language capabilities of an LLM. However, SuperICL requires fine-tuning the SLM on the target dataset, which assumes heavy compute resources and a large amount of high-quality data. This cancels the main benefit of ICL which is its low cost. Additionally fine-tuning may lead to data leakage issues.


\textbf{Contributions.} We propose Ensemble SuperICL, an ICL method that enables an LLM to leverage the predictions and confidence scores of several off-the-shelf SLMs. We test Ensemble SuperICL on several natural language understanding benchmarks and show that it outperforms ICL, SLM, and SuperICL baselines. To demonstrate the transferability of expertise across tasks, we test Ensemble SuperICL on a medical labelling task, and show that our method can label large-scale domain-specific data more accurately than all baselines. We conduct an ablation study and sensitivity analyses to validate results.

\section{Ensemble Super In-Context Learning}

\begin{figure}[t!]
\begin{center}
\centerline{\includegraphics[width=\textwidth]{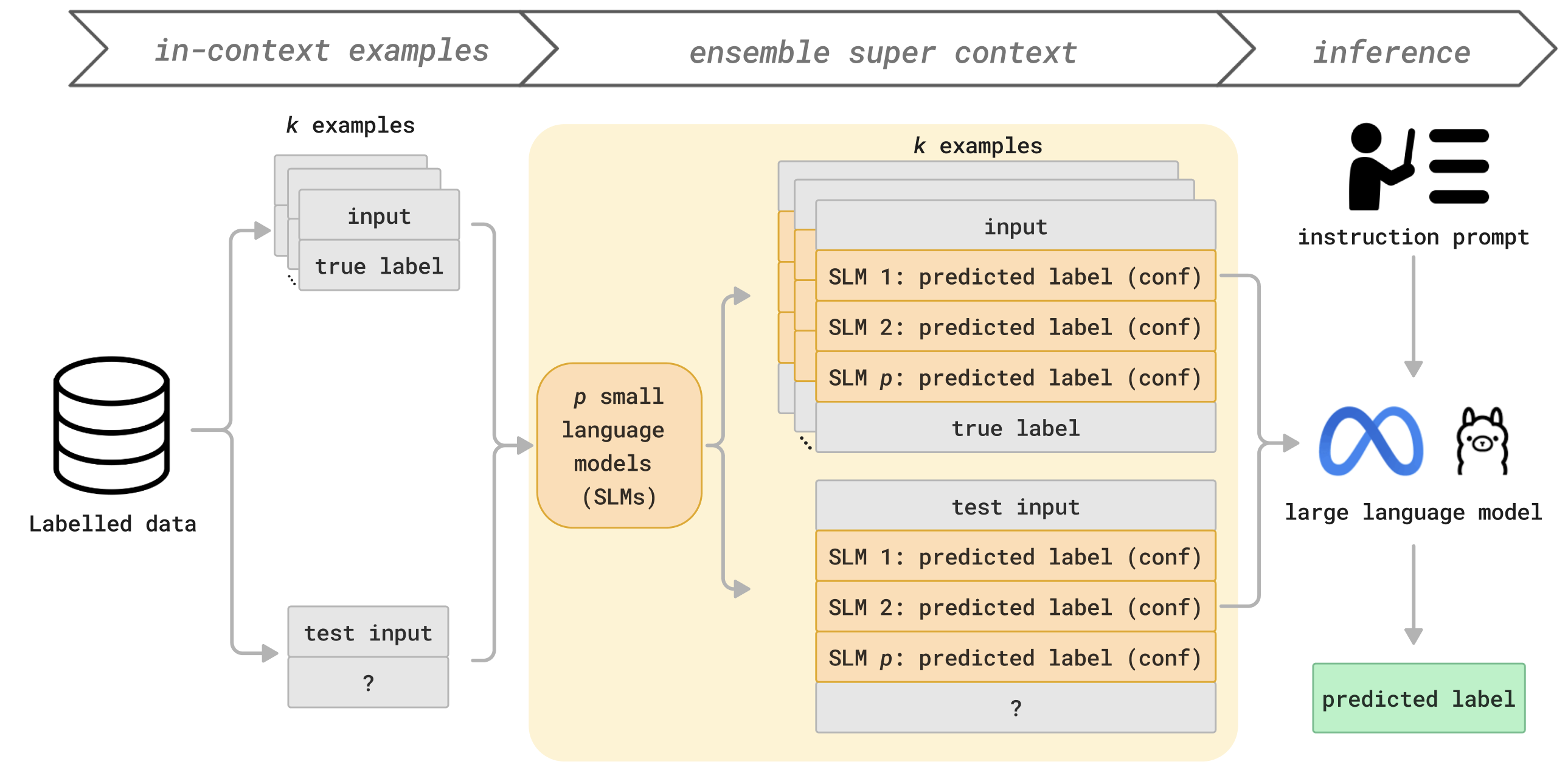}}
\caption[Schematic of Ensemble SuperICL]{\textbf{Three stages in Ensemble SuperICL.} \textit{In-context examples}: $k$ input-true label pairs are sampled from a dataset, along with a test input. \textit{Ensemble super context}: the examples are passed to $p$ small language models (SLMs) which each produce predicted labels and confidence scores, for the examples
and the test input. \textit{Inference}: a large language model (LLM) is given an instruction prompt along with the SLM-enhanced examples and test input, and must predict the label of the test input.}
\label{fig:schematic_esupericl}
\end{center}
\vskip -0.3in
\end{figure}


As their parameter counts scale to billions, transformer-based language models begin to exhibit in-context learning abilities which allow them to perform tasks they were not trained to do \citep{brown2020language}. Few-shot prompting leverages an LLM's ICL ability by providing the LLM with a few carefully chosen demonstrations (in-context examples) \citep{dong2022survey}.\footnote{As the dominant form of in-context learning (ICL), few-shot prompting is often used synonymously with ICL, as is the case in this work.}


Past works have explored both ensembling in-context demonstrations and incorporating SLMs in ICL. \citet{khalifa2023exploring} proposed demonstration ensembling, where an LLM splits demonstrations into subsets, produces output probabilities for each subset, and combines the probabilities for a final prediction. \citet{sun2023text} used a SLM to perform \textit{k}-Nearest Neighbours demonstration search for ICL. \citet{hugginggpt2024} used an LLM as a controller to access many expert models, small and large, for complicated tasks. \citet{wang2024large} developed an algorithm that uses a small language model to select optimal demonstrations from a set of annotated data.

Most recently, \citet{supericl} and \citet{yang2023supervised} proposed SuperICL and SuperContext, both methods that enrich the in-context demonstrations typical of ICL with the predictions of a SLM as well as its confidence scores, represented as the sigmoid applied to the logit probabilities produced by the SLM for each prediction. The smaller models provide task-specific knowledge, acting as plug-in models to the LLM that possesses general language capabilities.

Building on these past works, we propose \textit{Ensemble SuperICL}, a method which leverages ICL to enable an LLM to draw on the expert knowledge of several SLMs. A flowchart of the Ensemble SuperICL process is shown in Figure \ref{fig:schematic_esupericl}.
The first step in Ensemble SuperICL is selecting in-context examples where each example is a pair of input and true label. Next, two or more fine-tuned SLMs (which are fine-tuned directly on the task or on a general task) produce \textit{ensemble super context}: one demonstration in Ensemble SuperICL consists of an input, the predicted labels and confidence scores of two to five SLMs on this in-context example, and the true label. Finally, a test question is concatenated with its predicted labels and confidence scores from SLMs and fed to the LLM. To summarise, the LLM uses the constructed context and test question, both enhanced with the predictions and confidence scores of several SLMs, to formulate a response. An example of Ensemble SuperICL on the SST-2 dataset in a 1-shot setting with two SLMs is shown in Appendix \ref{esupericl_example}.

The motivation behind such context construction is that the LLM will be able to use the SLM predictions, their confidence scores, and the true label to both triangulate the correct response and learn how reliable a given SLM's predictions (and confidence) are.

\section{Methods}

\begin{table}[t!]
\caption{\textbf{The seven small language models (SLMs) considered in our analysis, their sizes (parameter count), and their accuracy on four benchmark datasets and one case study dataset (MedMCQA).} For the benchmarks, a version of the SLM fine-tuned on the benchmark tasks was used for evaluation (e.g., to evaluate MRPC, we used a version of RoBERTa-large fine-tuned on the MRPC dataset). For MedMCQA, all SLMs used were fine-tuned on one of the benchmark tasks, MNLI. Dashes indicate where fine-tuned SLMs were unavailable for a dataset or not considered.
}
\label{tab:slms}
\begin{center}
\begin{tabular}{llcccccc}
\toprule
SLM           & Size & SST-2 & MRPC  & MNLI  & CoLA  & MedMCQA \\
\midrule
MobileBERT    & 25M  & -     & -     & -     & 52.78 & -      \\
flan-t5-base  & 248M & -     & -     & 88.68 & -     & 70.43  \\
ELECTRA-large & 335M & 96.56 & 89.95 & 90.28 & 67.43 & 29.86  \\
DeBERTa-large & 350M & 94.95 & 89.71 & 90.39 & 64.06 & 71.43  \\
RoBERTa-large & 356M & 96.44 & 89.71 & 88.68 & 65.65 & 61.57  \\
BART-large    & 407M & 95.30 & 87.50 & 88.85 & -     & 68.71  \\
T5-large      & 770M & -     & -     & -     & 53.51 & -      \\
\bottomrule
\end{tabular}
\end{center}
\vskip -0.1in
\vspace{-1.3em}
\end{table}

\textbf{Datasets.} We used five datasets for our experiments: four natural language understanding (NLU) benchmarks and one domain-specific dataset. The General Language Understanding Evaluation (GLUE) benchmark is a collection of resources for assessing NLU systems \citep{wang2019glue}. We used four of the eleven GLUE datasets to evaluate a range of NLU abilities: the Multi-Genre Natural Language Inference corpus (MNLI), the Stanford Sentiment Treebank (SST-2), the Microsoft Research Paraphrase Corpus (MRPC), and the Corpus of Linguistic Acceptability (CoLA) (Appendix \ref{datasets}) \citep{dolan2005mrpc, socher2013sst, williams2018mnli, warstadt2018cola}. In addition, the Medical Multiple-Choice Question Answering dataset (MedMCQA) contains over 183k medical entrance exam questions. Each question is assigned one of 21 medical subjects such as surgery, dental, and pathology. We task our models with inferring the subject of a given question.

\textbf{Ensemble SuperICL Models.} We used Llama3-8b-Instruct as the LLM, and considered seven SLMs: MobileBERT, flan-t5-large, ELECTRA-large, DeBERTa-large, RoBERTa-large, BART-large, and T5-large (Table \ref{tab:slms}). We used fine-tuned versions of the SLMs from Hugging Face for each dataset, except for MedMCQA where we reused the MNLI fine-tuned SLMs. MNLI is one of the most popular dataset choices for fine-tuning SLMs, providing them with potentially transferable general language understanding abilities. We ran experiments with 2, 3, 4, and 5 choices of SLMs, and considered all possible combinations of SLMs over 0, 8, 16, 24, and 32-shot settings.

\textbf{Baselines.} We consider three baselines. The first is traditional in-context learning with 8, 16, 24, and 32-shot examples, and the second is SuperICL, the original method that uses a single SLM. Since our method demands the use of several small language models (SLMs), we construct a third baseline predictor: a simple majority vote of the SLM predictions. If there is no majority vote, as is possible in cases where only four SLMs were considered, the last appearing prediction is used. After considering several versions of majority vote predictors, including one weighted by the SLM confidence scores, we found that this approach yielded the highest accuracy across datasets, on average. This is because underperforming SLMs can be confidently wrong.

\textbf{Constructing the context.} For each dataset, we randomly sample 8, 16, 24, or 32 question-answer pairs from the training set to use for context construction. For fair comparison and reproducibility, we use the same in-context demonstrations for a dataset across all experiments.

\textbf{Inference time.} The ensemble super context, test input, and an instruction prompt are given to the LLM to predict the label of the test input (see Appendix \ref{prompt_designs} for specific prompt designs). For fair comparison and reproducibility, the LLM in our analyses predicts the token with the highest probability (greedy decoding) rather than sampling from the output distribution. All experiments were run on two NVIDIA A100 80GB GPUs.

\section{Results and Discussion}

\subsection{Ensemble SuperICL outperforms nearly all baselines on each dataset}

A summary of the results can be seen in Table \ref{tab:main_results}. Ensemble SuperICL boosted ICL performance by 3 to 20 percentage points across datasets, with greater gains on more challenging tasks, and outperforms all baselines on three out of four benchmark datasets: SST-2, MRPC, and CoLA. On MNLI, Ensemble SuperICL outperforms the ICL and SuperICL baselines, but the SLM majority vote baseline is marginally better. Extensive results from over 500 experiments containing all few-shot settings and SLM combinations are reported in Appendices \ref{main_experiments} and \ref{medmcqa_experiments}.

Ensemble SuperICL outperforms all baselines on the MedMCQA labelling task. Note that the SLMs used were not fine-tuned on the target task. This suggests that ensembling SLMs (tuned on general reasoning tasks) can outperform ICL on a domain-specific task while preserving its low data and compute requirements.

\begin{table}[t!]
\caption{\textbf{Ensemble SuperICL (E-SuperICL) outperforms all baselines on three natural language understanding benchmarks and a domain-specific labelling task (MedMCQA).} We present accuracies of the best performing versions of each order of Ensemble SuperICL (2, 3, 4, and 5 SLMs) on four GLUE benchmark datasets and a medical domain labelling task based on the MedMCQA dataset. The baselines considered are ICL with Llama3-8b-Instruct, the original SuperICL which uses a single SLM \citep{supericl,yang2023supervised}, and a majority vote of the SLM predictions. Note: only four SLMs were considered in SST-2 and MRPC due to a lack of available fine-tuned versions.}
\label{tab:main_results}
\begin{center}
\begin{tabular}{lcccccc}
\toprule 
                        & SST-2 & MRPC  & MNLI  & CoLA  & MedMCQA \\
\midrule
ICL (Llama3-8b-Instruct)& 94.15 & 75.25 & 71.24 & 55.43 & 79.43 \\
SuperICL                & 96.56 & 89.22 & 87.45 & 67.21 & 82.71 \\
SLM Majority vote       & 96.67 & 90.69 & \textbf{91.39} & 65.64 & 68.14 \\
\midrule
E-SuperICL 2            & \textbf{97.13} & 90.69          & 88.41           & \textbf{70.36} & \textbf{84.29} \\
E-SuperICL 3            & 97.02          & \textbf{91.42} & 90.25           & \textbf{70.36} & 83.71 \\
E-SuperICL 4            & 96.79          & \textbf{91.42} & 90.47           & 70.32          & 82.43 \\
E-SuperICL 5            & -              & -              & 91.27           & 68.17          & 80.57 \\
\bottomrule

\end{tabular}
\end{center}
\vskip -0.2in
\end{table}

Surprisingly, even low-performing SLMs were useful in boosting ICL performance. In the experiments on CoLA, we included two `weak' SLMs that, despite being fine-tuned on the target task, performed worse than the LLM in ICL: MobileBERT and T5 performed with 52.78\% and 53.51\% accuracy, and Llama3-8b-Instruct ICL with 55.43\% accuracy, yet the combination of MobileBERT and T5 through Ensemble SuperICL outperformed ICL under all few-shot regimes, with 62.21\% accuracy at its highest (Table \ref{tab:main_results_cola} in Appendix \ref{main_experiments}). In MedMCQA, ELECTRA-large performed very poorly with 29.86\% accuracy, compared to Llama3-8b-Instruct ICL with 79.43\% accuracy, yet Ensemble SuperICL with ELECTRA-large outperformed ICL in three out of five few-shot regimes, with 82.00\% accuracy at its highest (Table \ref{tab:medmcqa_results} in Appendix \ref{medmcqa_experiments}). These results suggests that Ensemble SuperICL with `weak' SLMs may be sufficient to surpass ICL performance. We also demonstrate that our results remain valid across random seeds (Appendix \ref{sensitivity_examples}).

\subsection{Ablation study: all components of Ensemble SuperICL are significant}

We investigated the effects of three components on the best-performing versions of Ensemble SuperICL: (a) the SLM predictions for the in-context examples; (b) the confidence scores of the SLMs in both the in-context examples and test input; (c) the SLM predictions for the test input. The results of the ablation study are presented in Table \ref{tab:ablation}, suggesting all components are essential for optimal performance.

First (1), we retained only the SLM predictions in the test input, removing SLM predictions from in-context examples and the SLM confidence scores from the test input. On average, this has the most significant impact on performance, most notably in MNLI and MedMCQA where performance drops about 6 percentage points.
Second (2), we retained SLM predictions in both the in-context examples and test input, but removed SLM confidence scores in both. This has the most significant impact on performance in SST-2, MRPC, and CoLA, but has the least impact on MNLI and MedMCQA.
Third (3), we retained SLM predictions and  confidence scores in the test input, but removed both from the in-context examples. On average, this has the least significant impact on performance.


\begin{table}[t!]
\caption{\textbf{Ablation study on Ensemble SuperICL.} We run ablation experiments on the best performing versions of Ensemble SuperICL for each dataset, focusing on three components. (a) Ctxt is the SLM predictions for the in-context examples; (b) Conf is the SLM confidence scores in both the in-context examples and the test input; (c) Test represents the SLM predictions for the test input. Note: since the best-performing Ensemble SuperICL on MNLI is zero-shot, we use the second best version (24-shot) in this table to enable ablations on the in-context examples.}
\vskip 0.1in
\label{tab:ablation}
\centering
\begin{tabular}{lcccccccc}
\toprule
& \multicolumn{3}{c}{Components} & \multicolumn{4}{c}{} \\
\cmidrule(lr){2-4}
 & (a) Ctxt & (b) Conf & (c) Test & SST-2 & MRPC & MNLI & CoLA & MedMCQA \\
\midrule
(1) & $\times$ & $\times$ & $\checkmark$         & 96.90 & 89.71 & 84.76 & 67.94 & 78.57 \\
(2) & $\checkmark$ & $\times$ & $\checkmark$     & 96.79 & 89.71 & 89.41 & 67.25 & 83.00 \\
(3) & $\times$ & $\checkmark$ & $\checkmark$     & 97.02 & \textbf{91.42} & 86.94 & 69.85 & 78.86 \\
\midrule
(4) & $\checkmark$ & $\checkmark$ & $\checkmark$ & \textbf{97.13} & \textbf{91.42} & \textbf{91.14} & \textbf{70.36} & \textbf{84.29} \\
\bottomrule
\end{tabular}
\vspace{-0.6em}
\end{table}

\subsection{Limitations and future works}

First, the biggest limitation of Ensemble SuperICL is that the optimal configuration of hyperparameters, such as number of in-context examples, number of SLMs, and choice of SLMs, varies considerably across datasets (Appendix \ref{hyperparameters}). However, Ensemble SuperICL outperforms ICL regardless of hyperparameters, on average, suggesting it may still be useful without a judicious hyperparameter selection process. Further, the scope of this work was limited to text classification tasks. \citet{yang2023supervised} showed positive results for SuperICL on text generation tasks, which suggests Ensemble SuperICL should generalise to text generation as well. A follow-up study may therefore test Ensemble SuperICL on question answering datasets such as SQuAD 2.0, QNLI, and QQP \citep{squad2018,wang2019glue,qqp2018}. In the same vein, a future work may expand beyond the medical domain to better evaluate the domain specialisation capabilities of Ensemble SuperICL.

\section{Conclusion}

We present Ensemble SuperICL, a method that ensembles small language models (SLMs) to improve ICL classification accuracy while preserving its low time, compute, and data requirements. Ensemble SuperICL outperforms all baselines on three natural language understanding (NLU) benchmarks, as well as a medical classification task, showing promise in generalising to domain-specific classification tasks. Two results further support its case for real-world applications: (1) the `harder' the task for an LLM (as measured by ICL accuracy), the more Ensemble SuperICL boosts ICL performance, and (2) even the combination of weak SLMs may be sufficient to obtain high performance. Result (2) maintains the `cheap' nature of ICL while benefiting from the gains of using small language models, and result (1) makes the method attractive for labelling tasks that require domain expertise.

\section*{Acknowledgements}
L.Y. was supported by EPSRC [EP/S026347/1] and the Hong Kong Innovation and Technology Commission (InnoHK Project CIMDA). A.M. was partially supported by the Oxford Internet Institute’s Research Programme funded by the Dieter Schwarz Stiftung gGmbH.

\bibliographystyle{plainnat}
\bibliography{refs}

\newpage
\appendix

\section{Datasets}
\label{datasets}

We used five datasets for our experiments, four of which are popular benchmark sets from the General Language Understanding Evaluation benchmark (GLUE).

\begin{table}[H]
\centering
\renewcommand{\arraystretch}{1.3}
\caption{Datasets, tasks, number of classes, metrics, and the number of evaluation examples for each dataset. All are classification tasks. Matthew's correlation coefficient is a popular metric for unbalanced datasets; it produces a high score only if the prediction obtained good results in all of the four confusion matrix categories (true positives, false negatives, true negatives, and false positives).}
\vskip 0.1in
\begin{tabular}{lcccc}
\hline
\textbf{Dataset}                  & \textbf{Task}                              & \textbf{\# Classes} & \textbf{Metric} & \textbf{\# Eval} \\
\hline

Glue-SST2 \citep{socher2013sst}    & \makecell{sentiment\\prediction}           & 2 & accuracy        & 872 \\
Glue-MRPC \citep{dolan2005mrpc}    & \makecell{paraphrase\\detection}           & 2 & accuracy        & 408 \\
Glue-MNLI \citep{williams2018mnli} & \makecell{natural language\\inference}     & 3 & accuracy        & 9815 \\
Glue-CoLA \citep{warstadt2018cola} & \makecell{grammatical\\acceptability}      & 2 & \makecell{matthew's\\correlation\\coefficient} & 1043 \\
MedMCQA        \citep{medmcqa2022} & \makecell{medical subject\\classification} & 2 & accuracy        & 700 \\

\hline
\end{tabular}
\label{tab:datasets}
\end{table}






For the domain-specific case study, we used the Medical Multiple-Choice Question Answering dataset (MedMCQA), a corpus of over 183k medical entrance exam questions \citep{medmcqa2022}. Each question in MedMCQA is assigned one of 21 medical subjects such as Surgery, Dental, and Pathology. We created a randomly sampled, balanced dataset of 700 questions labelled as either Dental or Surgery (to maximise amount of test data).
One example of the questions is given below:
\begin{description}
    \item[\hspace{2em}Question:] Till what age tetracycline should not be given to prevent discoloration?
    \item[\hspace{2em}Label:] Dental
\end{description}

\section{Example of Ensemble SuperICL procedure on the SST-2 dataset}
\label{esupericl_example}

\begin{figure}[h!]
\begin{center}
\centerline{\includegraphics[width=0.67\columnwidth]{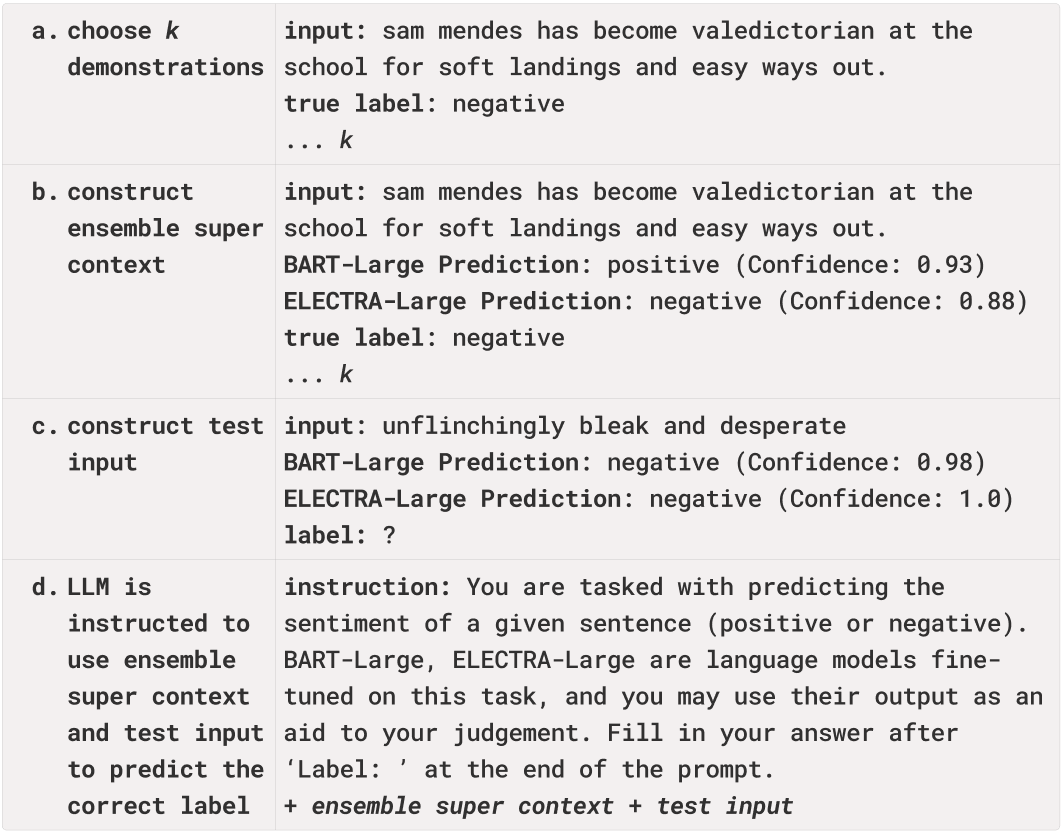}}
\caption{\textbf{Example of Ensemble SuperICL procedure on the SST-2 dataset.}}
\label{fig:esupericl_example}
\end{center}
\vskip -0.2in
\end{figure}

Figure \ref{fig:esupericl_example} gives a detailed example of the steps of Ensemble SuperICL applied to the SST-2 dataset.

\section{Prompt designs for each dataset}
\label{prompt_designs}
This section presents the detailed prompt designs for each dataset:
\vskip 0.2cm
\textbf{MNLI:}
\begin{verbatim}
    `role': `system',
        `content': You are tasked with determining the relationship between
        a pair of sentences as entailment (the hypothesis is a true conclusion
        from the premise), contradiction (the hypothesis contradicts the
        premise), or neutral (the hypothesis neither necessarily follows from
        nor contradicts the premise). RoBERTa-Large is a language model fine-
        tuned on this task, and you may use its output as an aid to your
        judgement. Fill in your answer after `Label: ' at the end of the
        prompt. [in-context examples]'
    `role': `user',
        `content': `[test example]
                     Label: ',
\end{verbatim}

\vskip 0.2cm
\textbf{SST-2:}
\begin{verbatim}
    `role': `system',
        `content': You are tasked with predicting the sentiment of a given
        sentence (positive or negative). RoBERTa-Large is a language model
        fine-tuned on this task, and you may use its output as an aid to your
        judgement. Fill in your answer after `Label: ' at the end of the
        prompt. [in-context examples]'
    `role': `user',
        `content': `[test example]
                     Label: ',
\end{verbatim}

\vskip 0.2cm
\textbf{MRPC:}
\begin{verbatim}
    `role': `system',
        `content': You are tasked with determining whether two sentences
        are semantically equivalent (equivalent or not_equivalent).
        RoBERTa-Large is a language model fine-tuned on this task, and
        you may use its output as an aid to your judgement. Fill in your
        answer after `Label: ' at the end of the prompt. [in-context examples]'
    `role': `user',
        `content': `[test example]
                     Label: ',
\end{verbatim}

\vskip 0.2cm
\textbf{CoLA:}
\begin{verbatim}
    `role': `system',
        `content': You are tasked with determining whether the grammar
        of a given sentence is correct (acceptable or unacceptable).
        RoBERTa-Large is a language model fine-tuned on this task, and
        you may use its output as an aid to your judgement. Fill in
        your answer after `Label: ' at the end of the prompt. [in-context
        examples]'
    `role': `user',
        `content': `[test example]
                     Label: ',
\end{verbatim}

\vskip 1.5cm
\textbf{MedMCQA:}
\begin{verbatim}
    `role': `system',
        `content': You are tasked with determining the medical subject
        that a given question belongs to (Dental or Surgery). RoBERTa-Large
        is a language model fine-tuned on this task, and you may use its
        output as an aid to your judgement. Fill in your answer after
        `Label: ' at the end of the prompt. [in-context examples]'
    `role': `user',
        `content': `[test example]
                     Label: ',
\end{verbatim}

\section{Full results of experiments on benchmark datasets}
\label{main_experiments}

This section reports the results for all experiments conducted on the benchmark datasets where the small language models (SLMs) used were fine-tuned on the dataset being tested. 

\begin{table}[H]
\centering
\caption{\textbf{Ensemble SuperICL outperforms all baselines, except SLM majority vote on the MNLI-m dataset.} Results of all orders of Ensemble SuperICL with Llama3-8b-Instruct and up to 5 SLMs on three datasets from the General Language Understanding Evaluation benchmark (GLUE). The first two row groups contain baseline comparisons: in-context learning (ICL) where the LLM can only reference demonstrations, the performance of each fine-tuned small language model (SLM) in a zero-shot setting,
and the performance of a majority vote classifier with all SLMs. The majority vote classifier is defined as the mode of the set of SLM predictions.}
\label{tab:main_results_full}
\vskip 0.1in
\resizebox{\columnwidth}{!}{%
\begin{tabular}{l p{.670cm}p{.670cm}p{.670cm}p{.670cm}p{.670cm}p{.670cm}p{.670cm}p{.670cm}p{.670cm}p{.670cm}p{.670cm}p{.670cm}p{.670cm}p{.670cm}p{.670cm}p{.670cm}}
  \toprule
  & \multicolumn{5}{c}{MNLI-m} & \multicolumn{5}{c}{SST-2} & \multicolumn{5}{c}{MRPC}\\
       \cmidrule(lr){2-6}\cmidrule(lr){7-11}\cmidrule(lr){12-16}
  Model & 0-shot & 8-shot & 16-shot & 24-shot & 32-shot & 0-shot & 8-shot & 16-shot & 24-shot & 32-shot & 0-shot & 8-shot & 16-shot & 24-shot & 32-shot \\
  \midrule
   ICL (Llama3-8b-Instruct) & 63.62 & 68.63 & 69.48 & \textbf{71.24} & 67.90 & 89.45 & 90.83 & 93.81 & 94.04 & \textbf{94.15} & 67.16 & 66.91 & 66.18 & 71.57 & \textbf{75.25} \\  
   RoBERTa       & 88.68 & - & - & - & - & 96.44 & - & - & - & - & 89.71 & - & - & - & -\\
   DeBERTa       & 90.39 & - & - & - & - & 94.95 & - & - & - & - & 89.71 & - & - & - & -\\
   BART          & 88.85 & - & - & - & - & 95.30 & - & - & - & - & 87.50 & - & - & - & -\\
   flan-t5       & 88.68 & - & - & - & - &   -   & - & - & - & - &   -   & - & - & - & -\\
   ELECTRA       & 90.28 & - & - & - & - & 96.56 & - & - & - & - & 89.95 & - & - & - & -\\
   Majority vote & 91.39 & - & - & - & - & 96.67 & - & - & - & - & 90.69 & - & - & - & -\\
   \midrule
   R & 86.71 & 83.48 & 80.48 & 81.78 & 78.59 & 94.95 & 96.10 & 96.45 & 96.22 & 96.22 & 89.71 & 87.50 & 88.48 & 87.01 & 87.26 \\
   D & 88.26 & 84.70 & 83.50 & 83.74 & 79.53 & 94.04 & 95.07 & 95.30 & 95.76 & 95.76 & 89.71 & 87.50 & 88.24 & 86.28 & 86.52 \\
   B & 87.75 & 85.23 & 82.23 & 82.69 & 80.26 & 94.73 & 95.41 & 95.76 & 96.10 & 96.22 & 87.75 & 86.28 & 86.52 & 85.78 & 84.56 \\
   f & 85.60 & 83.50 & 80.79 & 81.48 & 78.15 & -     & - & - & - & - & - & - & - & - & - \\
   E & \textbf{88.42} & 87.45 & 85.09 & 86.38 & 81.38 & 95.41 & 96.45 & 96.33 & \textbf{96.56} & 96.45 & \textbf{89.95} & 88.73 & 89.22 & 88.48 & 87.01 \\
   \midrule
   R+D & 90.70 & 84.66 & 82.61 & 85.81 & 83.78 & 95.64 & 95.64 & 95.99 & 96.56 & 96.67 & 89.71 & 89.95 & 89.71 & 89.22 & 87.50 \\
   R+B & 89.96 & 84.86 & 83.79 & 86.07 & 83.93 & 95.76 & 96.10 & 96.10 & 96.67 & 96.45 & 90.44 & 89.71 & 88.97 & 88.97 & 86.03 \\
   R+f & 89.15 & 86.00 & 84.20 & 85.62 & 83.40 & - & - & - & - & - & - & - & - & - & - \\
   R+E & 90.67 & 85.20 & 84.74 & 87.33 & 84.98 & 96.33 & 96.33 & 96.67 & 97.02 & \textbf{97.13} & 90.20 & 90.44 & 89.95 & 89.95 & 88.73 \\
   D+B & 90.60 & 85.79 & 84.30 & 86.38 & 84.14 & 95.76 & 95.99 & 95.87 & 95.87 & 95.87 & 90.20 & 89.95 & 88.97 & 86.77 & 83.82 \\
   D+f & 90.25 & 87.40 & 85.78 & 87.12 & 84.60 & - & - & - & - & - & - & - & - & - & - \\
   D+E & \textbf{91.01} & 86.59 & 86.02 & 88.09 & 85.85 & 96.33 & 96.10 & 96.33 & 96.67 & 96.79 & 90.20 & 90.69 & 90.20 & 89.22 & 88.97 \\
   B+f & 89.32 & 86.72 & 83.75 & 86.12 & 82.07 & - & - & - & - & - & - & - & - & - & - \\
   B+E & 90.59 & 87.68 & 87.13 & 88.41 & 86.48 & 96.22 & 95.87 & 96.10 & 96.90 & 96.79 & \textbf{90.93} & 90.44 & 90.44 & 89.71 & 88.73 \\
   f+E & 89.94 & 87.83 & 86.93 & 88.15 & 85.90 & - & - & - & - & - & - & - & - & - & - \\
   \midrule
   R+D+B   & 90.78 & 85.79 & 86.29 & 89.36 & 87.63 & 96.22 &  95.76 & 95.76 & 95.99 &  96.10 & 91.18 & 90.44 & 90.93 & 89.46 & 89.46 \\
   R+D+f   & 90.36 & 87.83 & 87.56 & 89.81 & 88.24 &  - & - & - & - & - & - & - & - & - & - \\
   R+D+E   & 91.14 & 86.69 & 86.64 & 89.29 & 87.55 & 96.67 &  95.99 & 96.67 & \textbf{97.02} &  96.90 & 90.93 & \textbf{91.42} & 90.93 & 90.93 & 90.20 \\
   R+B+f   & 89.89 & 87.23 & 87.44 & 88.88 & 87.29 &  - & - & - & - & - & - & - & - & - & - \\
   R+B+E   & 90.82 & 86.75 & 87.09 & 89.26 & 86.79 & 96.33 &  96.01 & 96.56& 96.79&  96.56 & 91.18 & 90.93 & 91.18 & 90.44 & 89.46 \\
   R+f+E   & 90.69 & 88.67 & 87.34 & 89.30 & 88.30 &  - & - & - & - & - & - & - & - & - & - \\
   D+B+f   & 90.69 & 87.63 & 87.79 & 89.36 & 87.56 &  - & - & - & - & - & - & - & - & - & - \\
   D+B+E   & \textbf{91.21} & 88.15 & 87.87 & 89.32 & 87.87 & 95.87 &  95.99 & 96.45 & 96.67 & 96.79 & 90.44 & 90.69 & 90.44 & 88.24 & 89.22 \\
   D+f+E   & 90.97 & 89.60 & 88.24 & 89.37 & 88.97 &  - & - & - & - & - & - & - & - & - & - \\
   B+f+E   & 90.82 & 89.58 & 89.05 & 90.25 & 89.48 &  - & - & - & - & - & - & - & - & - & - \\
  \midrule
   R+D+B+f & 90.54 & 88.47 & 90.02 & 89.65 & 89.80 & - &  - & - & - & - & - & - & - & - & - \\
   R+D+B+E & \textbf{91.10} & 88.60 & 89.66 & 89.79 & 89.90 & 96.33 & 95.99 & 96.45 & \textbf{96.79} & 96.56 & \textbf{91.42} & \textbf{91.42} & 90.69 & 90.93 & 90.20 \\
   R+D+f+E & 91.03 & 88.74 & 89.99 & 90.23 & 89.77 & - &  - & - & - & - & - & - & - & - & - \\
   D+B+f+E & 91.04 & 89.52 & 90.47 & 90.34 & 90.04 &  - & - & - & - & - & - & - & - & - & - \\
   R+B+f+E & 90.76 & 89.55 & 90.39 & 90.39 & 89.95 &  - & - & - & - & - & - & - & - & - & - \\
     \midrule
   R+D+B+f+E & \textbf{91.27} &  90.40 & 90.96 & 91.14 & 90.89 & - & - & - & - & - & - & - & - & - & - \\
  \bottomrule
\end{tabular}
}

\end{table}

\thispagestyle{empty} 
\begin{table}[H]
\centering
\caption{\textbf{Ensemble SuperICL outperforms all baselines on the Corpus of Linguistic Acceptability dataset (CoLA).} Results of all orders of Ensemble SuperICL with Llama3-8b-Instruct and up to 5 SLMs on CoLA. The first two row groups contain baseline comparisons: in-context learning (ICL) where the LLM can only use demonstrations as reference, the performance of each fine-tuned small language model (SLM) in a zero-shot setting,
and the performance of a majority vote classifier with all SLMs. The majority vote classifier is defined as the mode of the set of SLM predictions. The metric reported in this table is Matthew's Correlation.}
\label{tab:main_results_cola}
\begin{tabular}{l p{.670cm}p{.670cm}p{.670cm}p{.670cm}p{.670cm}p{.670cm}}
  \toprule
  & \multicolumn{5}{c}{CoLA}\\
       \cmidrule(lr){2-6}
  Model & 0-shot & 8-shot & 16-shot & 24-shot & 32-shot \\
  \midrule
   ICL (Llama3-8b-Instruct) & 52.16 & 55.19 & \textbf{55.43} & 52.52 & 54.90 \\  
   RoBERTa (R)    & 65.65 & - & - & - & - \\
   DeBERTa (D)    & 64.06 & - & - & - & - \\
   MobileBERT (M) & 52.78 & - & - & - & - \\
   ELECTRA (E)    & 67.43 & - & - & - & - \\
   T5 (T)         & 53.51 & - & - & - & - \\
   Majority vote  & 65.64 & - & - & - & - \\
   \midrule
   R & 65.92 & 65.67 & 65.67 & 65.05 & 65.00\\
   D & 64.56 & 63.83 & 63.60 & 64.61 & 64.58\\
   M & 54.70 & 60.33 & 57.39 & 57.67 & 56.76\\
   E & \textbf{67.70} & 67.21 & 67.21 & 66.76 & 67.01\\
   T & 55.99 & 57.84 & 56.18 & 57.00 & 55.29\\
   \midrule
   R+D & 68.96 & 68.47 & 67.71 & 69.25 & 67.44\\
   R+M & 66.06 & 64.49 & 63.35 & 64.82 & 64.31\\
   R+E & 69.76 & 70.26 & \textbf{70.36} & 69.34 & 69.87\\
   R+T & 66.19 & 66.29 & 64.56 & 67.46 & 66.03 \\
   D+M & 64.13 & 64.47 & 66.14 & 66.36 & 65.92\\
   D+E & 68.05 & 68.25 & 69.09 & 68.80 & 69.57 \\
   D+T & 65.32 & 65.17 & 67.10 & 68.25 & 67.14 \\
   M+E & 66.17 & 69.20 & 67.88 & 67.04 & 68.18\\
   M+T & 58.85 & 62.21 & 59.24 & 59.95 & 59.00\\
   E+T & 67.80 & 65.77 & 66.11 & 66.67 & 69.28 \\
   \midrule
   R+D+M   & 67.51 & 66.67 & 64.78 & 65.28 & 64.52\\
   R+D+E   & 69.98 & 68.47 & 68.74 & 69.24 & 69.83 \\
   R+D+T   & 68.37 & 64.67 & 65.11 & 66.45 & 67.31\\
   R+M+E   & 67.44 & 66.67 & 68.63 & \textbf{70.36} & 69.78\\
   R+M+T   & 67.34 & 66.14 & 65.83 & 66.45 & 67.17\\
   R+E+T   & 68.78 & 64.61 & 65.33 & 66.94 & 67.42 \\
   D+M+E   & 68.33 & 66.93 & 67.82 & 68.41 & 68.59\\
   D+M+T   & 65.80 & 62.52 & 65.67 & 66.70 & 67.21\\
   D+E+T   & 68.28 & 64.30 & 66.57 & 67.16 & 69.80\\
   M+E+T   & 64.11 & 64.73 & 65.56 & 67.20 & 66.95\\
  \midrule
   R+D+M+E & 69.52 & 69.02 & 68.20 & 68.86 & 67.50\\
   R+D+M+T & 68.59 & 64.78 & 66.16 & 66.82 & 66.52\\
   R+D+E+T & 68.88 & 68.41 & \textbf{70.32} & 67.37 & 67.27\\
   D+M+E+T & 69.45 & 65.58 & 65.57 & 68.28 & 67.73\\
   R+M+E+T & 68.17 & 65.67 & 66.91 & 67.91 & 65.54\\
  \midrule
   R+D+M+E+T & \textbf{69.44} & 63.92 & 68.17 & 67.49 & 66.27\\
  \bottomrule
\end{tabular}

\end{table}

\newpage
\section{Full results of experiments from MedMCQA case study}
\label{medmcqa_experiments}
Recall we randomly sampled 350 questions labelled as both surgery and dental for a total of 700 questions for our experiments. Given a question from these 700, the LLM is tasked with assigning the correct subject (dental or surgery). To show the practicality of our method, we deploy Ensemble SuperICL using small language models (SLMs) fine-tuned on MNLI, a general natural language understanding task.


\begin{table}[H]
\centering
\caption{\textbf{Ensemble SuperICL outperforms all baselines on the MedMCQA labelling task, without the need for the small language models to be fine-tuned on medical domain tasks.} Results of all orders of Ensemble SuperICL on the MedMCQA labelling task with Llama3-8b-Instruct and \textit{up to 5 SLMs, all fine-tuned on MNLI}. The first two row groups contain baseline comparisons: in-context learning (ICL) where the LLM can only reference demonstrations, the performance of each fine-tuned small language model (SLM) in a zero-shot setting, and the performance of a majority vote classifier with all SLMs. The majority vote classifier is defined as the mode of the set of SLM predictions.}
\label{tab:medmcqa_results}
\scalebox{0.9}{
\begin{tabular}{l p{.670cm}p{.670cm}p{.670cm}p{.670cm}p{.670cm}}
  \toprule
  & \multicolumn{5}{c}{MedMCQA} \\
       \cmidrule(lr){2-6}
  Model & 0-shot & 8-shot & 16-shot & 24-shot & 32-shot \\
  \midrule
   ICL (Llama3-8b-Instruct) & 70.29 & 75.29 & 74.14 & 78.00 & \textbf{79.43} \\
   RoBERTa (MNLI-tuned)     & 61.57 & - & - & - & - \\
   DeBERTa (MNLI-tuned)     & 71.43 & - & - & - & - \\
   BART (MNLI-tuned)        & 68.71 & - & - & - & - \\
   flan-t5 (MNLI-tuned)     & 70.43 & - & - & - & - \\
   ELECTRA (MNLI-tuned)     & 29.86 & - & - & - & - \\
   Majority vote            & 68.14 & - & - & - & - \\
   \midrule
   R & 71.43 & 82.29 & 78.57 & \textbf{82.71} & 80.00 \\
   D & 74.43 & 82.00 & \textbf{82.71} & \textbf{82.71} & 82.29 \\
   B & 72.00 & 80.29 & 80.43 & 80.00 & 81.14 \\
   f & 75.43 & 78.71 & 81.71 & 79.71 & 82.00 \\
   E & 57.00 & 78.57 & 79.43 & 82.00 & 80.00 \\
   \midrule
   R+D & 77.57 & 79.29 & 81.71 & 82.57 & 82.71 \\
   R+B & 76.57 & 82.14 & 80.29 & 82.43 & 80.57 \\
   R+f & 76.86 & 83.29 & 83.00 & 82.14 & 82.00 \\
   R+E & 70.14 & 78.86 & 82.29 & 81.57 & 82.57 \\
   D+B & 78.29 & 80.00 & 80.71 & 81.57 & 80.14 \\
   D+f & 79.71 & 80.71 & \textbf{84.29} & 82.29 & 82.57 \\
   D+E & 72.57 & 81.57 & 81.71 & 81.29 & 79.71 \\
   B+f & 78.00 & 81.57 & 82.86 & 81.43 & 82.14 \\
   B+E & 71.29 & 78.29 & 77.43 & 79.14 & 79.71 \\
   f+E & 74.29 & 79.00 & 81.43 & 81.57 & 81.71 \\
   \midrule
   R+D+B   & 78.00 & 82.71 & 81.43 & 81.43 & 80.29 \\
   R+D+f   & 78.86 & 80.57 & 82.86 & 81.00 & 80.00 \\
   R+D+E   & 75.57 & 82.00 & 79.71 & 83.00 & 80.14 \\
   R+B+f   & 77.86 & 80.14 & 83.14 & 81.57 & 78.86 \\
   R+B+E   & 74.86 & 78.86 & 79.57 & 82.29 & 80.00 \\
   R+f+E   & 74.86 & 82.29 & 82.29 & \textbf{83.71} & 80.71 \\
   D+B+f   & 79.00 & 79.71 & 81.00 & 82.00 & 79.29 \\
   D+B+E   & 74.14 & 79.14 & 79.43 & 80.14 & 79.00 \\
   D+f+E   & 77.00 & 81.29 & 80.29 & 82.57 & 80.14 \\
   B+f+E   & 75.00 & 80.57 & 80.57 & 80.86 & 79.00 \\
  \midrule
   R+D+B+f & 78.71 & \textbf{82.43} & 81.43 & 81.57 & 79.14 \\
   R+D+B+E & 74.43 & 81.71 & 80.14 & 81.43 & 79.43 \\
   R+D+f+E & 77.43 & 81.57 & 79.86 & 82.00 & 80.29 \\
   D+B+f+E & 77.00 & 80.00 & 80.43 & 81.43 & 78.57 \\
   R+B+f+E & 76.43 & 81.29 & 79.43 & 81.71 & 80.57 \\
     \midrule
   R+D+B+f+E & 76.43 & 79.29 & 80.29 & \textbf{80.57} & 79.43 \\
  \bottomrule
\end{tabular}
}

\end{table}

\section{Best hyperparameters of Ensemble SuperICL}
\label{hyperparameters}

Table \ref{tab:esupericl_specs} gives the optimal set of Ensemble SuperICL hyperparameters for each dataset.

\begin{table}[H]
\caption{The optimal hyperparameters for Ensemble SuperICL can vary across datasets.}
\label{tab:esupericl_specs}
\vskip 0.1in
\begin{center}
\begin{tabular}{lcccc}
\toprule 
                        & \makecell{\# of few-shot\\examples}  & \# of SLMs & \makecell{choice of\\base SLMs} & \makecell{SLMs\\fine-tuned on} \\
\midrule
SST-2                   & 32    & 2     & \makecell{RoBERTa-Large,\\ELECTRA-Large}                                             & SST-2   \\
\midrule
MRPC                    & 8     & 3,4   & \makecell{RoBERTa-Large,\\DeBERTa-Large,\\BART-Large,\\ELECTRA-Large}                & MRPC    \\
\midrule
MNLI                  & 0     & 5     & \makecell{RoBERTa-Large,\\DeBERTa-Large,\\BART-Large,\\flan-t5-base,\\ELECTRA-Large} & MNLI  \\
\midrule
CoLA                    & 16,24 & 2,3   & \makecell{RoBERTA-Large,\\MobileBERT,\\ELECTRA-Large}                                & CoLA    \\
\midrule
MedMCQA                 & 16    & 2     & \makecell{DeBERTa-Large,\\flan-t5-base}                                              & MNLI \\
\bottomrule
\end{tabular}
\end{center}
\vskip -0.1in
\end{table}

\section{Variance in Ensemble SuperICL is generally low}
\label{sensitivity_examples}

Table \ref{tab:sensitivity} shows that our results are robust to choosing different in-context examples.


\begin{table}[h]
\caption{\textbf{Sensitivity of Ensemble SuperICL to varying in-context examples}. The best performing versions of Ensemble SuperICL for each dataset with different seeds for in-context examples. 
Var = variance. Note: since the best-performing Ensemble SuperICL on MNLI is zero-shot, the second best version (24-shot) was used in this table to enable ablations on the in-context examples.}
\label{tab:sensitivity}
\centering
\begin{tabular}{llccccc|c}
\toprule
& & \multicolumn{5}{c}{Random seed} & Var \\
    \cmidrule(lr){3-7}
     & & 42 & 0 & 1 & 2 & 3 & \\
    \midrule
    \multirow{3}{*}{\rotatebox[origin=c]{90}{SST-2}}
    &ICL & 94.15 & 93.81 & 94.27 & 93.69 & 95.07 & 0.29 \\
    &SLM MV & 96.67 & 96.67 & 96.67 & 96.67 & 96.67 & - \\
    &E-SuperICL & \textbf{97.13} & \textbf{97.02} & \textbf{96.79} & \textbf{97.02} & \textbf{96.56} & 0.05 \\
    \midrule
    \multirow{3}{*}{\rotatebox[origin=c]{90}{MRPC}}
    &ICL & 75.25 & 75.49 & 74.51 & 74.27 & 73.78 & 0.50 \\
    &SLM MV & 90.69 & 90.69 & 90.69 & 90.69 & 90.69 & - \\
    &E-SuperICL & \textbf{91.42} & \textbf{89.95} & \textbf{91.18} & \textbf{88.73} & \textbf{90.69} & 1.18 \\
    \midrule
    \multirow{3}{*}{\rotatebox[origin=c]{90}{MNLI}}
    &ICL & 71.24 & 62.28 & 71.37 & 70.10 & 70.13 & 14.57 \\
    &SLM MV & \textbf{91.39} & \textbf{91.39} & \textbf{91.39} & \textbf{91.39} & 91.39 & - \\
    &E-SuperICL & 91.14 & 91.35 & 91.21 & 91.13 & \textbf{91.50} & 0.02 \\
    \midrule
    \multirow{3}{*}{\rotatebox[origin=c]{90}{CoLA}}
    &ICL & 55.43 & 54.50 & 54.58 & 53.71 & 55.07 & 0.42 \\
    &SLM MV & 65.64 & 65.64 & 65.64 & 65.64 & 65.64 & - \\
    &E-SuperICL & \textbf{70.36} & \textbf{70.25} & \textbf{69.73} & \textbf{69.00} & \textbf{68.20} & 0.82 \\
    \midrule
    \multirow{3}{*}{\rotatebox[origin=c]{90}{\makecell{Med-\\MCQA}}}
    &ICL & 79.43 & 73.86 & 80.00 & 77.57 & 75.57 & 6.68 \\
    &SLM MV & 68.14 & 68.14 & 68.14 & 68.14 & 68.14 & - \\
    &E-SuperICL & \textbf{84.29} & \textbf{81.43} & \textbf{82.00} & \textbf{80.00} & \textbf{82.43} & 2.44 \\
\bottomrule
\end{tabular}
\end{table}


\end{document}